\documentclass[sigconf]{acmart}

\AtBeginDocument{%
  \providecommand\BibTeX{{%
    \normalfont B\kern-0.5em{\scshape i\kern-0.25em b}\kern-0.8em\TeX}}}

\usepackage{amsmath}
\usepackage[title]{appendix}
\usepackage{makecell}
\usepackage{graphicx}
\usepackage{subfig}
\usepackage{multirow}

\usepackage{natbib}
\copyrightyear{2021} 
\acmYear{2021} 
\setcopyright{othergov}\acmConference[ICCPS '21]{ACM/IEEE 12th International Conference on Cyber-Physical Systems (with CPS-IoT Week 2021)}{May 19--21, 2021}{Nashville, TN, USA}
\acmBooktitle{ACM/IEEE 12th International Conference on Cyber-Physical Systems (with CPS-IoT Week 2021) (ICCPS '21), May 19--21, 2021, Nashville, TN, USA}
\acmPrice{15.00}
\acmDOI{10.1145/3450267.3452000}
\acmISBN{978-1-4503-8353-0/21/05}
\author{Yeli Feng}
\affiliation{%
  \institution{Nanyang Technological University}
}
\author{Arvind Easwaran}
\affiliation{%
  \institution{Nanyang Technological University}
}

\newcommand{\ignore}[1]{}

\begin{document}


\copyrightyear{2021}

\title{WiP Abstract : Robust Out-of-distribution Motion Detection and Localization in Autonomous CPS }

\begin{abstract}
Highly complex deep learning models are increasingly integrated into modern cyber-physical systems (CPS), many of which have strict safety requirements. One problem arising from this is that deep learning lacks interpretability, operating as a black box. The reliability of deep learning is heavily impacted by how well the model training data represents runtime test data, especially when the input space dimension is high as natural images. In response, we propose a robust out-of-distribution (OOD) detection framework. Our approach detects unusual movements from driving video in real-time by combining classical optic flow operation with representation learning via variational autoencoder (VAE). We also design a method to locate OOD factors in images. Evaluation on a driving simulation data set shows that our approach is statistically more robust than related works.  
\end{abstract}

\maketitle

\section{INTRODUCTION} \label{section:1}
With state-of-the-art deep neural networks (DNN), close to perfectly accurate object classifiers can be trained given large amounts of data. But at runtime, it is well known that these classifiers' prediction accuracy tends to reduce when novel data outside the training data distribution is encountered. When the input space dimension is extremely high, it is impossible to obtain a complete combination of images and ground truths for model training and verification. For the same reason, whether the distribution of training data accurately represents the true probability distribution of the input space is hard to be verified. It is a challenge to integrate many useful or promising machine learning functions into safety-critical CPS. For example, in recent years, self-driving cars caused several fatalities under rare driving circumstances, including unusual maneuvers from vehicles around and the occurrence of unexpected pedestrians. We define images including rare factors with regards to training data as out-of-distribution (OOD) samples. They are perceptual variations in images caused by lighting and weather conditions, viewing angles, degradation in image quality due to damage in sensors, new objects in the world, and many more. 

Detection of OOD samples or anomalies is an active research topic. One research direction exploits the statistics exhibited in the neural units of SoftMax layers in well-trained DNN, i.e., the controller, as shown in figure ~\ref{fig:fig1}. Different distance metrics are proposed to separate OOD samples from in-distribution (ID) data \ignore{\cite{jiang2018trust}}. The generative adversarial network is also explored. In CPS, the OOD detection problem tends to be addressed along a line independent of the controller.  Several related works design VAE-based methods to detect OOD samples from input or latent spaces \cite{cai:iccps2020, wip:2020}. In the autonomous driving setting, detecting potential hazardous situations such as night and rain is studied widely. 

Besides visibility factors, another key to road safety is that learning-enabled autonomous CPS remains vigilant to external motion patterns distinct from those previously learned, thus preventing a driving model from acting on unreliable maneuver decisions. Driving conditions are immensely diverse. A car crash could happen on the highway, in downtown, or residential areas. Hence the input videos include very complex semantics and in high dimension. We propose a 2-tier representation abstraction approach to tackle the challenges. Our work contributes distinctly in:

\textbf{\textit{Robustness}} Related works  \cite{cai:iccps2020, wip:2020} rely only on VAE's information bottleneck to address above mentioned two challenges. We introduce optic flow operation to reduce the semantic complexity in the VAE input space dramatically. A comparison study with related works shows our method is significantly more robust. Our OOD detector achieves the highest accuracy while maintains the lowest false position rate. 

\textbf{\textit{Localization}} We design a neural activation-based method to locate OOD factors detected in images to facilitate reasoning, which is not addressed in related works.

\section{PROPOSED FRAMEWORK} \label{section:2}
\vspace{-3mm}
\begin{figure}[h]
  \centering
  \includegraphics[width=0.8\linewidth]{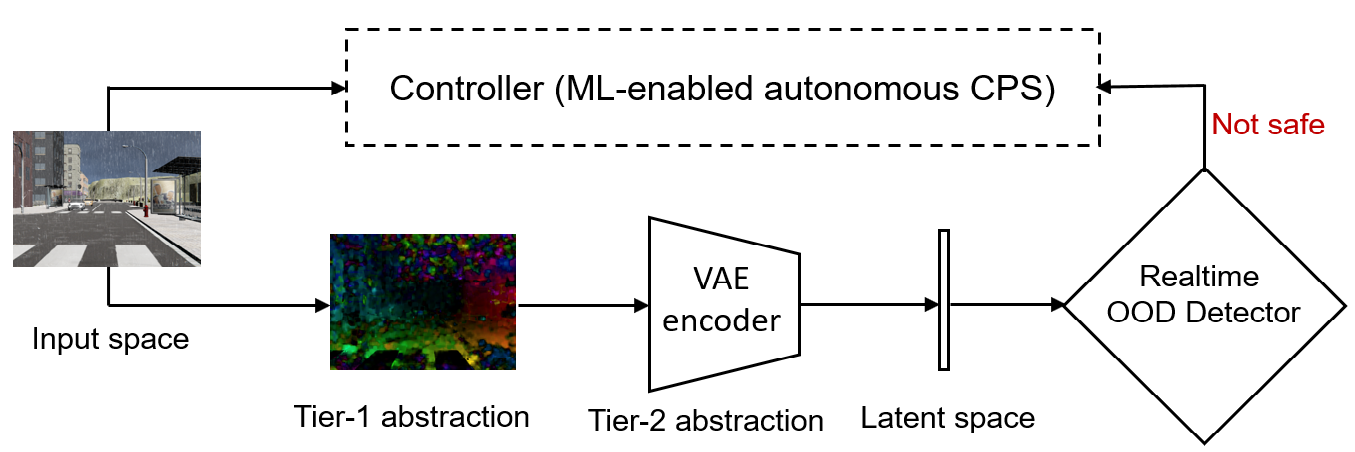}
  \vspace{-2mm}
  \caption{Schematic Diagram of OOD Detection Proposal}
  \label{fig:fig1}
\end{figure}
\vspace{-3mm}
As illustrated in figure ~\ref{fig:fig1}, in the first tier of representation abstraction, optic flow matrices are computed from the input video's neighboring image frames. Assuming intensities of pixels in a region remain constant across neighboring frames, movement velocity can be solved by the optic flow constraint equation \ignore{\cite{lucas1981iterative}}. The outputs here are 2D arrays in the same dimension of input images. It represents environmental motions from which we want to recognize patterns distinct from those in training data. For example, an autonomous driving system might never encounter a situation where a big object suddenly cuts into its driving lane. We encode such an abrupt driving lane cut as an OOD factor, manifested in limitless semantic combinations in our world. 

Existing approaches in the literature learn OOD factors from color images. For fast-moving cars to safely avoid collision or crash, the semantic meanings coded in pixel colors are less informative and could even be interpreted as semantic noise. We exploit optic flow operation to remove such noise, leading to more robust detection of motion OOD factors.  

Let \(x\) be training data with ID motions only. We design a convolutional VAE network to encode optic flow outputs into a 1D vector of much lower dimension, namely the latent variables \(z\). The training objective of VAE is to minimize the kullback–leibler (KL) divergence between the approximated posterior distribution \(\phi(z \mid x)\) of latent variables and true prior \(p(z)\) and the negative likelihood of reconstruction \(p(x \mid z)\) in the input space. Equation (\ref{eq:1}) defines our OOD detection measure. \(m\) is the dimension of latent variables \(z\). A properly trained VAE encoder will return close to zero \(D_{KL}\) value if given ID samples but high positive values over OOD samples.

To enhance real-time detection reliability, we design an inductive conformal anomaly detection (ICAD) based OOD detector. Instead of making a classify decision over a single test image, the OOD detector takes \(D_{KL}\) values of an image sequence as nonconformity scores, computes their p-values against a calibration set. A mixture martingale \(M_n\) is then calculated using these p-values. A calibration set consists of ID samples hold out from VAE model training. See in equations \ref{eq:2}. \(l\) and \(n\) are the numbers of samples in the calibration set and a test sequence, respectively.
\vspace{-2mm}
\begin{equation}
\label{eq:1}
\alpha = \sum_{i=1}^m D_{kl}(q(z \mid x) \parallel p(z))
\end{equation}
\vspace{-4mm}
\begin{equation}
\label{eq:2}
p_{l+1}= \frac{ \bigl\vert \; {i=1,...l} \mid \alpha_{i} \geqslant \alpha_{l+1} \; \bigl\vert } { (l+i)}
\;\;\;\;\;\;\;\;\;\;\;\;
M_n = \int\limits_0^1 \prod_{i=1}^n \varepsilon p_i^{\varepsilon-1} \ d\varepsilon
\end{equation}

\vspace{-1mm}
In \cite{zhou2016learning}, a class activation maps (CAM) method is proposed to explain why objects are recognized from images. We design a CAM and ICAD inspired intuitive way to locate OOD factors detected. A couple of OOD localization is shown in the top row of figure ~\ref{fig:fig2}. The red overlay indicates where the OOD is. The overlay map represents the relative activation strengthen between a test sample and the calibration set. The output from the last convolutional layer of the VAE encoder is used here. The OOD localization serves as a mechanism to identify the source of high nonconformity scores for reasoning purposes. 

\section{DETECTION EVALUATION} \label{section:3}
To compare detection performance with related work, we use one VAE architecture for all detection methods to evaluate, i.e., four CNN layers of 32/64/128/256 for both encoder and decoder, except the dimensions of latent spaces are different. A size of 24 is used in our method. SYNTHIA-AL \cite{bengar2019temporal}, a large-scale driving simulation data set, is utilized for model training and performance evaluation.

The evaluation set has 300 video episodes, each of 60 frames long. OOD factors include rain, car crashes, driving lane cuts by spawned car or pedestrian, dangerous tailgating, and turns at a road intersection.  The mixture martingale \(M_n\) value can be huge, so \(log(M_n)\) is used here. \(log(M_n)\) values from each detection method vary, grid parameter searches are carried out to find a detection threshold for each. \(log(M_n)=3\) is optimal for \cite{cai:iccps2020} and our detectors, whereas 10 is optimal for \cite{wip:2020}.  The number of frames \(n\) has few impacts on performance, so \(n=10\) for all methods. 

\begin{figure}[h]
  \centering
  \includegraphics[width=0.8\linewidth]{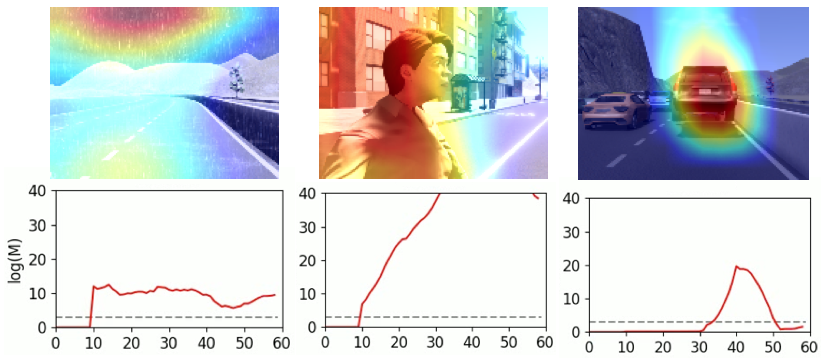}
   \vspace{-4mm} 
  \caption{OOD Detection and Localization }
  \vspace{-2mm}  
  \label{fig:fig2}
\end{figure}

Martingale curves of 3 test episodes are shown in the bottom row of figure \ref{fig:fig2}, where the dashed lines are the detection threshold. When the red curve is above the dashed line 10 frames or more, an OOD instance is detected. From the summary in table ~\ref{table:1}, we can observe that the proposed detection method achieves the highest true positive rate (TPR), lowest false positive rate (FPR), and tops the F1 score. Hence it is very robust.
\vspace{-2mm}
\begin{table}[h]
\caption{ Performance Comparison of OOD Detectors } 
\vspace{-2.5mm}
\label{table:1}
{
  	\begin{tabular}{ccccl}
    	\toprule
    	Detector & F1 score & Accuracy & TPR & FPR\\
    	\midrule
    	\cite{cai:iccps2020} & 0.773 & 0.673 & 0.884 & 0.685 \\  
     	\cite{wip:2020} & 0.877 & 0.830 & 0.958 & 0.387 \\   
     	Our  & \textbf{0.960} & \textbf{0.950} & 0.958 & \textbf{0.063} \\    
  	\bottomrule
	\end{tabular}
}
\end{table}
\vspace{-6mm}

\vspace{-1.5mm}
\textbf{\textit{Computational Cost}} Our method takes 126 milliseconds to make a detection decision on an Intel Xeon 3.50GHz CPU, while  \cite{cai:iccps2020} requires 127 and \cite{wip:2020} requires 71.Because of the optic flow operation, our method takes a longer time than \cite{wip:2020}. All three methods are efficient for real-time deployments.  

\section{CONCLUSION} \label{section:4}
This paper proposes a robust OOD detection framework that increases robustness. We also design an intuitive OOD localization mechanism to explain the detection results. We are interested in validating the proposed method over more diving data sets and more types of OOD factors and improving the framework in future work.
\vspace{-1mm}
\begin{acks}
This research was funded in part by MoE, Singapore, Tier-2 grant number MOE2019-T2-2-040.
\end{acks}
\vspace{-2mm}

\bibliography{ICCPS2021WiP}

\end{document}